# THAI RHETORICAL STRUCTURE ANALYSIS

Somnuk Sinthupoun
Department of Computer Science
Maejo University
Chiangmai, Thailand 50290
somnuk@mju.ac.th

Ohm Sornil
Department of Computer Science
National Institute of Development Administration
Bangkok, Thailand 10240
osornil@as.nida.ac.th

*Abstract*— **Rhetorical structure analysis (RSA) explores discourse relations among elementary discourse units (EDUs) in a text. It is very useful in many text processing tasks employing relationships among EDUs such as text understanding, summarization, and question-answering. Thai language with its distinctive linguistic characteristics requires a unique technique.**

**This article proposes an approach for Thai rhetorical structure analysis. First, EDUs are segmented by two hidden Markov models derived from syntactic rules. A rhetorical structure tree is constructed from a clustering technique with its similarity measure derived from Thai semantic rules. Then, a decision tree whose features derived from the semantic rules is used to determine discourse relations.**

I. INTRODUCTION

Rhetorical structure analysis (RSA) studies relations among elementary discourse units (EDUs). It provides a framework for analyses of text and is very useful to many text processing tasks employing relationships among EDUs such as text understanding, summarization, and question-answering.

Definition of EDU may vary. Some researchers consider an EDU to be a clause or a clause-like [6] excerpt while others consider them to be a sentence [14] in discourse parsing. A number of techniques are proposed to determine EDU boundaries for English language such as those using discourse cues [5, 12, 13], punctuation marks [6, 13], and syntactic information [6, 14, 15].

EDUs and their discourse relations (DRs) are commonly represented as a rhetorical structure tree (RS tree). It can be defined as follows: RS tree = (status, DR, promotion, left, right) where status is a set of EDUs; DR is a set of discourse relations; promotion is a subset of EDUs; and left and right can either be NULL or recursively defined objects of type RS tree [4, 6].

Many discourse relations can be used in writings. Some have a single nucleus such as elaboration and condition while others have multiple nucleuses such as contrast [25].

Marcu, et al. [7] determine discourse relations using Naive Bayes classifiers to learn all adjacent sentence pairs that contain the cue phrase (i.e. "but", "however") at the beginning of the second sentence, in the middle of a sentence, and at the end of the sentence. Pitler, et al. [9] determine local discourse relations using an N-gram model to compute transitional probabilities in both directions for each pair of EDUs. To account for remaining ambiguities, a unigram model based on previous known relations is used to predict the next one. Pitler, et al. [10] determine implicit discourse relations using naive Bayes, maximum entropy, and AdaBoost classifiers whose features include polarity tag, inquirer tag, verb classes, First-Last, First3, Modality, context and lexical features based on Penn Discourse Treebank [18].

For Thai language, Sukvaree, et al. [21] construct RS trees by using global and local spanning trees which determine relations by using DR marker tags. Wattanamethanont, et al. [15] purpose a technique to determine relations by using naive Bayes classifier whose features consist of DR marker, key phrase, and word co-occurrences.

This article proposes a new approach to Thai RSA which consists of three major steps: EDU segmentation, RS tree construction, and DR determination. Two hidden Markov models constructed from syntactic properties of Thai language are used in segmenting EDUs, a clustering technique with its similarity measure derived from semantic properties of Thai language is used to construct an RS tree, and a decision tree is used to determine the relation between two related EDUs in the RS tree.

II. ISSUES IN THAI RHETORICAL STRUCTURE ANALYSIS

Thai language has unique characteristics both syntactically and semantically. This makes techniques proposed for other languages not directly applicable to Thai language. A number of important issues with respect to Thai RSA are discussed in this section.

*A. No Explicit EDU Boundaries*

Unlike English, Thai language has no punctuation marks (e.g., comma, full stop, semi-colon, and blank) to determine the boundaries of EDUs. Therefore, EDU segmentation in Thai language becomes a nontrivial issue.

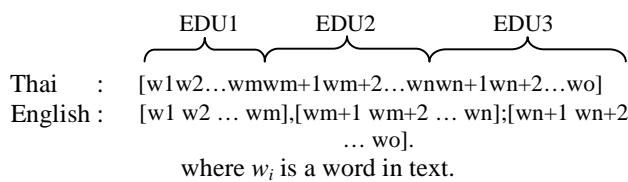

Thai : [w1w2…wmwm+1wm+2…wnwn+1wn+2…wo]
English : [w1 w2 … wm],[wm+1 wm+2 … wn];[wn+1 wn+2 … wo].
where $w_i$ is a word in text.

### B. EDU Constituent Omissions

Given two EDUs, an absence of subject, object or conjunction in the anaphoric EDU may happen, such as a situation where an anaphoric EDU omits the subject that refers back to the object of the cataphoric EDU. Accordingly, EDU boundaries are ambiguous.

Thai text : "เพื่อนจะขอยืมหนังสือ เพราะหาซื้อไม่ได้" (A friend's going to borrow this book because she hasn't been able to find it.)

Three possibilities :
1) [S(เพื่อน)V(จะขอยืม)O(หนังสือ)]$_{EDU1}$ [because S(Φ) V(หาซื้อไม่ได้)]$_{EDU2}$
2) [S(เพื่อน)V(จะขอยืม)O(หนังสือ)]$_{EDU1}$ [because(Φ)S(Φ)V(หาซื้อไม่ได้)]$_{EDU2}$
3) [S(เพื่อน)V(จะขอยืม)O(Φ)]$_{EDU1}$ [because(Φ)S(หนังสือ)V(หาซื้อไม่ได้)]$_{EDU2}$

In addition, the absence of subject, object or preposition which is a modifier nucleus of VP especially in the anaphoric EDU makes the use of word co-occurrence alone not sufficient to determine the relation between EDU1 and EDU2. For example,

EDU1: ศาลได้มีคำสั่งให้แยกสินสมรส (A court has ordered partition of marriage properties.)
EDU2: Φ1 จะสั่งยกเลิกการแยก Φ2 ได้ (Φ1 can cancel the partition of Φ2.)

In the example, EDU2 omits subject "ศาล" (court) and object 'สินสมรส' (marriage properties). Therefore, word co-occurrence alone is not sufficient to determine this relation.

### C. Implicit Markers

The absences of discourse markers in Thai language are often occurred. In the example below, "แต่" (but) is a discourse marker which is omitted, but the relation between EDU1 and EDU2 is still able to determine.

EDU1: ศาลได้มีคำสั่งให้แยกสินสมรส (A court has ordered partition of marriage property.)
EDU2: Φ กริยาหรือสามีคัดค้าน (Φ a wife or a husband may contest.)

Therefore, considering markers or cue phrases alone is not sufficient to determine the relation between EDUs.

### D. Adjacent Markers

Given three EDUs with two markers, as shown in the example below, two RS Trees are possible.

EDU1: ศาลได้มีคำสั่งให้แยกสินสมรส (A court has ordered partition of marriage properties.)
EDU2: แต่ถ้ากริยาหรือสามีคัดค้าน (but if a wife or a husband contests,)
EDU3: ศาลจะสั่งยกเลิกการแยกได้ (the court can cancel the partition.)

The first possibility, EDU1 and EDU2 relate first by a discourse marker "แต่" (but), next (EDU1, EDU2) and EDU3 relate by a marker "ถ้า" (if). For the other possibility, EDU2 and EDU3 relate first by a marker "ถ้า" (if), next that between (EDU2, EDU3) and EDU1 relate by a marker "แต่" (but).

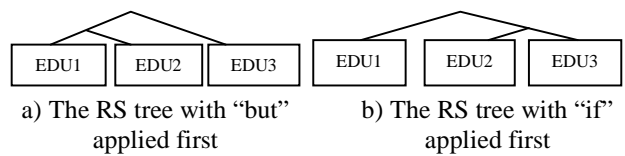

a) The RS tree with "but" applied first
b) The RS tree with "if" applied first

Fig. 1. Adjacent markers issue

### E. Marker Ambiguities

One marker may infer multiple relations such as "เมื่อ" (when) can infer condition or cause-result relation, and "แต่" (but) can infer "contrast" or "elaboration" relation whose example can be seen below:

EDU1: ศาลได้มีคำสั่งให้แยกสินสมรส (A court has ordered partition of marriage properties.)
EDU2: แต่กริยาหรือสามีคัดค้าน (but$_{contrast}$ a wife or a Husband may contest.)

On the other hand,
EDU1: ศาลได้มีคำสั่งให้แยกสินสมรส (A court has ordered partition of marriage properties.)
EDU2: แต่เฉพาะที่กริยาและสามีเห็นชอบ (but$_{elaboration}$ only what the wife and the husband agree.)

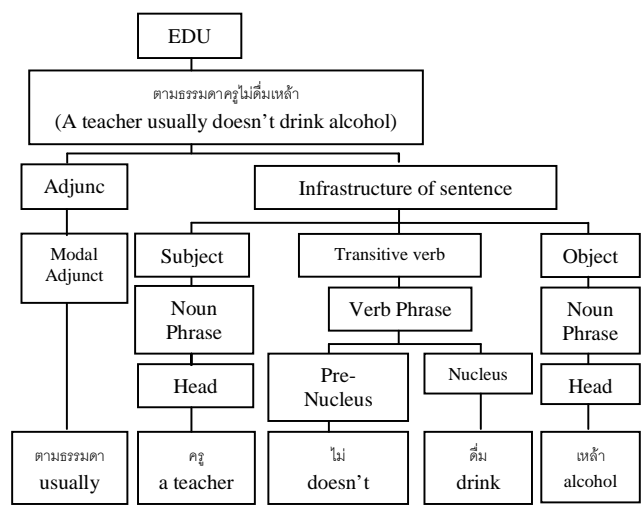

Fig. 2. Structure of the EDU "A teacher usually doesn't drink alcohol."

### III. STRUCTURES OF THAI EDUs

A Thai EDU consists of infrastructure and adjunct constituents. The twelve possible arrangements of Thai EDUs [22] are shown in Table 1. The structure of an EDU "A teacher usually doesn't drink alcohol" is shown in Fig. 2.

Table 1: The possible arrangements of Thai EDUs.

| EDUs | Examples | Rules |
|---|---|---|
| Vi | หิว (I'm hungry.) | $NP_S$-*Vi*-$NP_S$ |
| S-Vi | ฝน-ตก (It's rain.) | |
| Vi-S | เจ็บไหม-คุณ (Are you pain?) | |
| Vt-O | หิว- น้ำ (I'm hungry.) | $NP_O$-$NP_S$-*Vt*-$NP_O$ |
| S-Vt-O | รถ-ชน-เด็ก (The car hit the boy.) | |
| O-S-Vt | รูปนี้-ฉัน-ดูแล้วละจ๊ะ (I've already seen this photograph.) | |
| Vtt-O-I | ยังไม่ได้ให้-ยา-คนไข้ (I haven't given the patient the medicine.) | $NP_S$-*Vtt*-$NP_O$-$NP_I$ |
| S-Vtt-O-I | ใคร-ให้-ลูกกวาด-หนู (Who gave you the sweet?) | |
| O-S-Vtt-I | ความลับ-ใครละ-จะกล้าถาม- คุณ (Who would dare to ask you the secret?) | $NP_O$-$NP_S$-*Vtt*-$NP_I$ |
| I-S-Vtt-O | หนู-ป้า-จะให้-บ้านนี้ (Niece, I am going to give you this house.) | $NP_I$-$NP_S$-*Vtt*-$NP_O$ |
| N | ป้า (Auntie) | $NP_N$-$NP_N$ |
| N-N | นี่ปากกา-ใคร (Whose pen is this?) | |

### IV. EDU SEGMENTATION

This section describes the EDU segmentation technique proposed in this research. To reduce the segmentation ambiguities caused from omissions of words or discourse markers, and the appearances of modifiers, noun phrases and verb phrases which are constituents of EDUs are first determined, according to the syntactic properties of Thai language. These phrases are then used to identify boundaries of EDUs.

A noun phrase (NP) is a noun or a pronoun and its expansions which may function as one of the four Thai EDU constituents, namely subject (S), object (O), indirect object (Oi) and nomen (N). The general structure of a noun phrase consists of five constituents which are: head (H), intransitive modifier (Mi), adjunctive modifier (Ma), quantifier (Q), and determinative (D).

A verb phrase (VP) is a verb and its expansions which may function as one of the three Thai EDU constituents, namely intransitive verb (Vi), transitive verb (Vt) and double transitive verb (Vtt). The general structure of a verb phrase consists of four constituents which are: nucleus (Nuc), pre-nuclear auxiliary (Aux1), post-nuclear auxiliary (Aux2), and modifier (M).

There are twenty five possible arrangements of noun phrase and ten arrangements of verb phrases [22], which are shown in Table 2.

Table 2: The possible arrangements of Thai NPs and VPs.

| Noun Phrases | Noun Phrases (cont.) | Verb Phrases |
|---|---|---|
| H-Ma | H | Nuc |
| H-Mi-Ma | H-Mi | Nuc-Aux2 |
| H-Q-Ma | H-Q | Nuc-M |
| H-Ma-Q | H-D | Nuc-Aux2-M |
| H-D-Ma | H-Mi-Q | Nuc-M-Aux2 |
| H-Mi-Q-Ma | H-Q-Mi | Aux1-Nuc |
| H-Q-Mi-Ma | H-Mi-D | Aux1-Nuc-Aux2 |
| H-Mi-D-Ma | H-Q-D | Aux1-Nuc-M |
| H-Q-D-Ma | H-D-Q | Aux1-Nuc-Aux2-M |
| H-D-Q-Ma | H-Mi-Q-D | Aux1-Nuc-M-Aux2 |
| H-Mi-Q-D-Ma | H-Mi-D-Q | |
| H-Mi-D-Q-Ma | H-Q-Mi-D | |
| H-Q-Mi-D-Ma | | |

#### A. Phrase Identification

To perform phrase identification, word segmentation and part of speech (POS) tagging are performed using SWATH [20] which extracts words and classifies them into 44 types such as common noun (NCMN), active verb (VACT), personal pronoun (PPRS), definite determiner (DDAC), unit classifier (CNIT) and negate (NEG). A hidden Markov model (HMM) [18] employs these POS tag categories to determine phrases. The model assumes that at time step $t$ the system is in a hidden state $PC(t)$ which has a probability $b_{jk}$ of emitting a particular visible state of POS tag $tag(t)$, and a transition probability between hidden states $a_{ij}$:

$$a_{ij} = p(PC_j(t+1)/PC_i(t)). \quad (1)$$

$$b_{jk} = p(tag_k(t)/PC_j(t)). \quad (2)$$

where $PC(t)$ is the phrase constituent at time step $t$, and $tag(t)$ is POS tag at time step $t$.

The probability of a sequence of $T$ hidden states $PC^T = \{PC(1), PC(2), ..., PC(T)\}$ can be written as:

$$p(PC^T) = \prod_{t=1}^{T} p(PC(t) | PC(t-1)) \quad (3)$$

The probability that the model produces the corresponding sequence of POS tag $tag^T$, given a sequence of PCs $PC^T$ can be written as:

$$p(tag^T | PC^T) = \prod_{t=1}^{T} p(tag(t) | PC(t)) \quad (4)$$

Then, the probability that the model produces a sequence $tag^T$ of visible POS tag states is:

$$p(tag^T) = \arg\max_{PC_{1,n}} \prod_t^T p(tag(t)|PC(t)) p(PC(t)|PC(t-1)) \quad (5)$$

The Baum-Welch [18] learning algorithm is applied to determine model parameters, i.e., $a_{ij}$ and $b_{jk}$, from an ensemble of training samples.

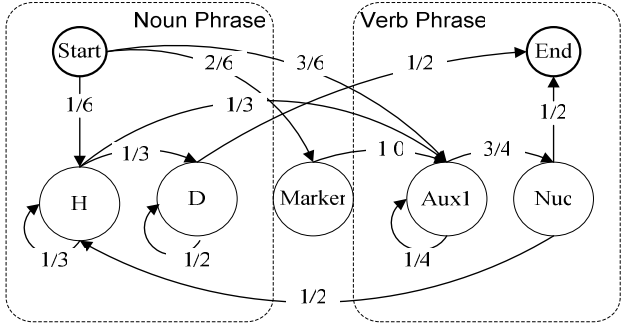

Fig. 3. A phrase identification model.

Given a sequence of visible state $tag^T$, the Viterbi algorithm [18] is used to find the most probable sequence of hidden states by recursively calculating $p(tag^T)$ of visible POS states. Each term $p(tag(t)/PC(t)) p(PC(t)/PC(t-1))$ involve only $tag(t)$, $PC(t)$, and $PC(t-1)$ by the following definition:

$$\delta_t(j) = \begin{cases} 0, & t=0 \text{ and } j \neq \text{initial state} \\ 1, & t=0 \text{ and } j = \text{initial state} \\ \arg\max_i \delta_{t-1}(i) a_{ij} b_{jkt}, & \text{otherwise} \end{cases} \quad (6)$$

where $b_{jkt}$ represents the transition probability $b_{jk}$ selected by the visible state emitted at time $t$. Thus, the only nonzero contribution to the arg is for index $k$ which matches the visible state $tag(t)$.

Figure 3 shows a phrase identification model of string "เพื่อนจะขอยืมหนังสือเล่มนี้ เพราะΦ₁ซื้อไม่ได้Φ₂ ดังนั้นΦ3จึงต้องยืมหนังสือฉัน" (A friend's going to borrow this book. Because she ($\Phi_1$) hasn't been able to buy it ($\Phi_2$). Therefore she ($\Phi_3$) must borrow it from me.) POS tags of the string is "เพื่อน (A friend-NCMN) จะขอ (is going to-XVMM) ยืม (borrow-VACT) หนังสือ (book-NCMN) เล่ม (numerative-CNIT) นี้ (this-DDAC) เพราะ (Because-CONJ) เธอ (she($\Phi_1$)-PPRS) ไม่ (hasn't been-NEG) สามารถ (able to-XVMM) ซื้อ (buy-VACT) มัน (it($\Phi_2$)) ดังนั้น (Therefore-CONJ) เธอ (she($\Phi_3$)-PPRS) จึงต้อง (must-XVMM) ยืม (borrow-VACT) หนังสือ (book-NCMM) ฉัน (me-PPRS)".

The hidden state of a phrase model consists of H(NCMN-book (2/4), -friend (1/4); PPRS-me (1/4)), D(CNIT-numerative (1/2); DDAC-this (1/2)), Discourse-marker(CONJ-because (1/2), -therefore (1/2)), Aux1(XVMM-is going to (1/4), -must (1/4), -able to (1/4); NEG-hasn't been (1/4)) and Nuc(VACT-borrow (2/3), -buy (1/3)).

*B. EDU Boundary Determination*

After we determine NPs and VPs, another HMM on EDU constituents (shown in Fig. 5.) is then created to determine the boundaries of EDUs. This model can handle the subject and object omission problems, discussed earlier.

Fig. 5 shows an example of the EDU segmentation model for an EDU "เพื่อน-จะขอ-ยืม-หนังสือ-เล่ม-นี้" (A friend's going to borrow this book.)

The EDU segmentation model can be expressed as:

$$p(tag^T) = \arg\max_{EDUC_{1,n}} \prod_t^T p(tag(t)|EDUC(t)) p(EDUC(t)|EDUC(t-1)) \quad (7)$$

where $EDUC(t)$ is EDU constituent at time step $t$, and $tag(t)$ is the phrase tag at time step $t$.

The expression, $p(EDUC(t)/EDUC(t-1))$ is the probability of EDU constituent (*EDUC*) at time $t$ given the previous *EDUC(t-1)*, and $p(tag(t)/EDUC(t))$ is the probability of phrase tag $tag(t)$ given $EDUC(t)$.

|  | เพื่อน | จะขอ | ยืม | หนังสือ | เล่ม | นี้ |  |
|---|---|---|---|---|---|---|---|
|  | Start | NCMN | XVMM | VACT | NCMN | CNIT | DDAC | END |
| Start | 1 | 0 | 0 | 0 | 0 | 0 | 0 | 0 |
| H | 0 | 1/6*3/4 | 0 | 0 | 8*10⁻³ | 0 | 0 | 0 |
| D | 0 | 0 | 0 | 0 | 0 | 1*10⁻³ | 3*10⁻⁴ | 0 |
| Marker | 0 | 2/6*0 | 0 | 0 | 0 | 0 | 0 | 0 |
| Aux1 | 0 | 3/6*0 | 3*10⁻² | 0 | 0 | 0 | 0 | 0 |
| Nuc | 0 | 0 | 0 | 2*10⁻² | 0 | 0 | 0 | 0 |
| End | 0 | 0 | 0 | 0 | 0 | 0 | 0 | 1*10⁻⁴ |
| T = | 0 | 1 | 2 | 3 | 4 | 5 | 6 | 7 |
| Output | Start | H | Aux1 | Nuc | H | D | D | End |

Fig.4. The results of Viterbi tagging on the phrase identification model in Fig.3.

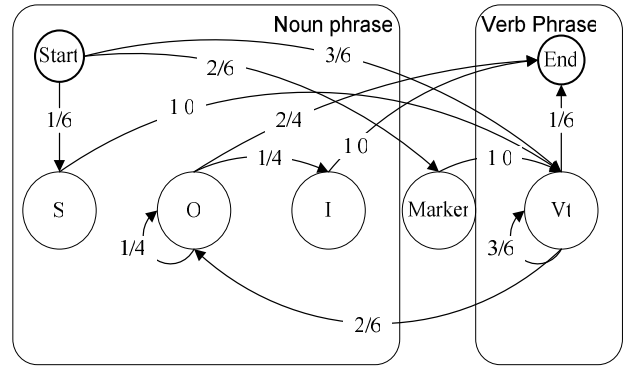

Fig.5. An example of a Thai EDU segmentation model.

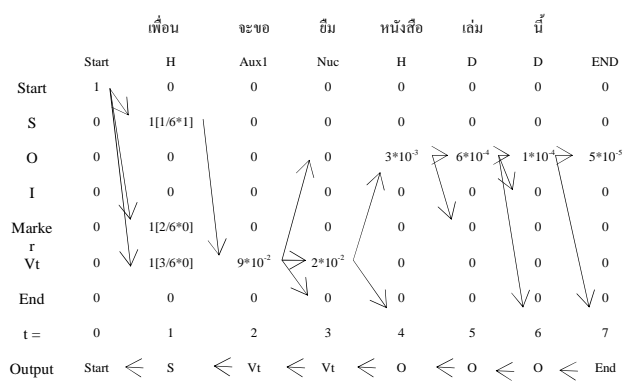

Fig.6. The results of Viterbi tagging on the Thai EDU segmentation model in Fig.5.

## C. EDU Constituent Grouping

Once EDU boundaries are determined, syntactic rules in Table 1 are then applied to group EDU constituents into a larger unit that will be used to match the semantic rules in further steps, For example a string "เพื่อน-จะขอ-ยืม-หนังสือ-เล่ม-นี้" (A friend's going to borrow this book.), the result from the Viterbi tagging on the EDU segmentation model is S, Vt, Vt, O, O, O. The matched rule of "$NP_O$-$NP_S$-Vt-$NP_O$" is applied, and the result becomes: "$NP_S – (V, V)_t – (NP, NP, NP)_O$."

## V. THAI RHETORICAL TREE CONSTRUCTION

In this section, we describe our proposed technique based on semantic rules derived from Thai linguistic characteristics to construct an RS tree from a corpus. The rules are classified into three types which are Absence, Repetition, and Addition rules [1, 3, 22, 23, 24]. Given a pair of EDUs, an author may write by using any combination of the rules. A similarity measure is calculated from these rules, and a hierarchical clustering algorithm employing this measure is used to construct an RS tree.

### A. Semantic Rules for EDU Relations

*1) Absence Rules*

In Thai language, it has been observed that frequently in writings some constituents of an EDU may be absent while its meaning remains the same. In the example below, the NP (object) "ขนม" (dessert) is absent from the anaphoric EDU, according to rule Φ (O, O).

Cataphoric EDU (Vt-O) : อยากจะทำขนมไหม (Would you like to make a dessert?)
Anaphoric EDU (Vt)   : อยากจะทำ (Yes, I do.)

*2) Repetition Rules*

It has been observed that frequently an anaphoric EDU relates to its cataphoric EDU by a repetition of NP (subject, object) or a preposition phrase (PP) functioning as a modifier of a nucleus or a verb phrase (VP). In the following example, two EDUs relate by a repetition of an object (NP) "บ้าน" (house), according to the rule я (O, O).

Cataphoric EDU (Vtt-O-I) : ผมกำลังจะขายบ้านให้เขา (I'm going to sell him a house.)
Anaphoric EDU (Vt-O)    : จะขายบ้านหลังไหน (Which house are you going to sell?)

*3) Addition Rules*

It has been observed that frequently an anaphoric EDU relates to its cataphoric EDU by an addition of a discourse marker, and possibly accompanied by Absence and/or Repetition rules. In the example below a discourse marker "เพราะ" (because) is added in front of the anaphoric EDU, according to the rule Д (Marker, Before).

Cataphoric EDU (Vtt-O-I) : ฉันอยากจะยืมหนัง (I want to borrow films.)
Anaphoric EDU (Vt-O)    : เพราะหาซื้อไม่ได้ (because I have not been able to buy it.)

Table 3 lists Repetition, Absence, and Addition rules, for example, я (S, S) means that the subject of the cataphoric EDU is repeated in the anaphoric EDU; Φ(S, S) means that the subject is present in the cataphoric EDU but absent from the anaphoric EDU; and Д (Marker, Before) means that a discourse marker is added in front of this particular EDU.

### B. EDU Similarity

Similarity between two EDUs can be calculated from the semantic rules in Table 3, as follows:

*1) Feature Calculations*

Given a pair of EDUs, for each rule, an EDU calculates a feature vector which consists of the following elements: Subject, Absence of Subject, Object, Absence of Object, Preposition, Absence of Preposition, Nucleus, Modifier Nucleus, Head, Absence of Head, Modifier Head, Absence of Modifier Head, Marker Before, and Marker After elements. The value of an element is dependent upon the type of rule, as follows:

Table 3: Repetition, Absence, and Addition rules.

| Repetition ( я ) | Absence ( Φ ) | Addition ( Д ) |
|---|---|---|
| я (S, S) | Φ (S, S) | Д (Marker, After) |
| я (O, S) | Φ (O, S) | Д (Marker, Before) |
| я (S, O) | Φ (S, O) | Д (Key Phrase, After) |
| я (O, O) | Φ (O, O) | Д (Key Phrase, Before) |
| я (S, Prep) | Φ (Only H, H) | |
| я (O, Prep) | Φ ((H, M), H) | |
| я (Prep, S) | Φ ((H, M), M) | |
| я (Prep, O) | Φ (S, Prep) | |
| я ((S, Prep), (S, Prep)) | Φ (O, Prep) | |
| я ((O, Prep), (S, Prep)) | Φ (Prep, S) | |
| я ((Prep, Prep), (S, Prep)) | Φ (Prep, O) | |
| я ((S, Prep), (O, Prep)) | | |
| я ((O, Prep), (O, Prep)) | | |
| я((Prep, Prep), (O, Prep)) | | |
| я (Only H, Only H) | | |
| я (H, M) | | |
| я (Only M, Only Nuc) | | |
| я (Only M, Only M) | | |
| я ((Nuc, M), (Nuc, M)) | | |

The following example is used to illustrate calculations related to semantic rules:

EDU1: ชาวบ้าน (Subject) ประกอบ (Nucleus) อุตสาหกรรมในครอบครัว (Object) (The villagers perform the family-industry.)

EDU2: และ (Before) Φ (Absence of Subject) หวงแหน (Nucleus) สมบัติของชาติ (Object) (and protect properties of the nation.)

EDU3: อุตสาหกรรมในครอบครัว (Subject) จึงเป็น (Nucleus) สมบัติของชาติ (Object) (Therefore, the family-industry is a property of the nation.)

To describe the calculations related to semantic rules, the following notations will be used. $C_{Cat}$ is a constituent of the cataphoric EDU, $C_{Ana}$ is a constituent of the anaphoric EDU, $Pos_{Cat}$ is the position of cataphoric EDU, and $Pos_{Ana}$ is the position of anaphoric EDU. $X:Y$ where $X$ can be either Cataphoric or Anaphoric, and $Y$ is an element in the vector of X, e.g., *Cataphoric:Subject* is the Subject element in the vector of the cataphoric EDU. $X:rule$ is an Addition rule applied to $X$ (i.e., a cataphoric or an anaphoric EDU).

*a) Features based on an Absence rules:*

Feature vectors of the cataphoric and anaphoric EDUs are filled for a matched Absence rule, as follows:

$$\text{If } \Phi(C_{Cat}, C_{Ana}) \text{ is true then}$$
$$\text{Cataphoric } C_{Cat} = \text{Anaphoric}(\text{Absence of } C_{Ana}) = 1 - \frac{|Pos_{Cat} - Pos_{Ana}|}{\text{Total \# of sentences}} \quad (8)$$

In this example, the properties of EDU1 and EDU2 match with the rule $\Phi(S, S)$ with the absence of subject "ชาวบ้าน" (villager) in the anaphoric EDU, thus:

$$\text{Cataphoric}: \text{Subject} = \text{Anaphoric}: \text{Absence of Subject} = 1 - \frac{|1-2|}{3} \quad (9)$$

*b) Features based on Repetition rules:*

Feature vectors of the cataphoric and anaphoric EDUs is filled for a matched Repetition rule, as follows:

$$\text{If } \Re(C_{Cat}, C_{Ana}) \text{ is true then}$$
$$\text{Cataphoric}: C_{Cat} = \text{Anaphoric}: C_{Ana}$$
$$= \frac{|Pos_{Cat} - Pos_{Ana}|}{\text{Total \# of sentences}} * \frac{\text{Total \# of repeating words}}{\text{Total \# of words in sentences}} \quad (10)$$

In the example, the properties of EDU1 and EDU3 match with the rule я (O, S) with a repetition of an object "อุตสาหกรรมในครอบครัว" (family-industries) in the cataphoric EDU as a subject in the anaphoric EDU, thus:

$$\text{Cataphoric}: \text{Object} = \text{Anaphoric}: \text{Subject} = (1 - \frac{1-3}{3}) * (\frac{1}{3} * \frac{1}{3}) \quad (11)$$

*c) Features based on Addition rules:*

Feature vectors of the cataphoric and anaphoric EDUs is filled for a matched Addition rule, as follows:

$$\text{If Cataphoric:} Д \text{ (Marker, After) is true then}$$
$$\text{Cataphoric:Marker After} = \text{Anaphoric:Marker Before} = 1$$
$$\text{else if Anaphoric:} Д \text{ (Marker, Before) is true then}$$
$$\text{Anaphoric:Marker Before} = \text{Cataphoric:Marker After} = 1 \quad (12)$$

In this example, the properties of EDU1 and EDU2 match with the rule Д (Marker, Before) at EDU2, thus:

$$\text{Anaphoric:Marker Before} = \text{Cataphoric:Marker After} = 1 \quad (13)$$

*2) Rule Scoring*

After for each rule, the two vectors of the EDU pair are calculated, the vectors are then combined into a rule score which depends on the type of rule and the distance between the two EDUs, as follows:

*a) Absence and Repetition Rules:*

These rules consist of two parts (cataphoric and anaphoric). If both parts of an Absence or a Repetition rule are true, then the rule is true. But if a part of an Absence or a Repetition rule is false, then the rule is false, thus:

$$\text{if } |Pos_{Cat} - Pos_{Ana}| < MD \text{ then}$$
$$RS_{\text{Absence or Repetition}} = [\text{Magnitude of } EDU_{Cataphoric} * \text{Magnitude of } EDU_{Anaphoric}] \quad (14)$$

where $Pos_{Cat}$ and $Pos_{Ana}$ are the positions of cataphoric and anaphoric EDUs, and *MD* is the maximum distance between the EDUs (from experiments *MD = 4* in this research)

*b) Addition Rules:*

In this type of rules, if one part of the rule is true, then the rule is true, thus:

$$\text{if } |Pos_{Cat} - Pos_{Ana}| < MD \text{ then}$$
$$RS_{\text{Addition}} = [\text{Magnitude of } EDU_{Cataphoric} + \text{Magnitude of } EDU_{Anaphoric}] \quad (15)$$

*3) Similarity Calculation*

Once rule scores are available, similarity between two EDUs (cataphoric and anaphoric) can be calculated as a sum of all the rule scores (each normalized into a range from 0 to 1) according to the CombSum method [8].

## C. Rhetorical Tree Construction

A hierarchical clustering algorithm is applied to create an RS tree where each sample (an EDU in this case) begins in a cluster of its own; and while there is more than one cluster left, two closest clusters are combined into a new cluster, and the distance between the newly formed cluster and each other cluster is calculated. Hierarchical clustering algorithms studied in this research are shown in Table 4, and two example RS trees created from two different algorithms are shown in Fig. 8.

Table 4. Hierarchical clustering algorithms studied in this research.

| Algorithms | Distance Between Two Clusters |
|---|---|
| Single Linkage | The smallest distance between a sample in cluster A and a sample in cluster B. |
| Unweighted Arithmetic Average | The average distance between a sample in cluster A and a sample in cluster B. |
| Neighbor Joining | A sample in cluster A and a sample in cluster B are the nearest. Therefore, define them as neighbors. |
| Weighted Arithmetic Average | The weighted average distance between a sample in cluster A and a sample in cluster B. |
| Minimum Variance | The increase in the mean squared deviation that would occur if clusters A and B were fused. |

## VI. DISCOURSE RELATION DETERMINATION

In this section, we describe our technique to determine relations based on features according to semantic rules in Table 3. A decision tree (C5.0 algorithm) employs these features to determine a relation.

### A. Feature Extaction

A feature score for discourse relation determination is calculated from contents of the EDUs, based on the three types of rules. The feature set consists of two subsets. The first subset is for the cataphoric EDU which consists of: Subject, Object, Preposition, Nucleus, Marker Before and Marker After. The other subset is for the anaphoric EDU which consists of: Subject, Absence of Subject, Object, Absence of Object, Preposition, Absence of Preposition, Nucleus, Modifier Nucleus, Head, Absence of Head, Modifier Head, Absence of Modifier Head, Marker Before, and Marker After. The value of each element is dependent upon the type of rules matched (multiple matching is allowed), as follows:

*1) Features based on Absence rules:*
Feature values are filled as follows:

If $\emptyset(C_{Cat}, C_{Ana})$ is true then
  $Cataphoric: C_{Cat} = 1$
  $Anaphoric: Absence\ of\ C_{Ana} = 1$ (16)

In the example, considering EDU1 and EDU2 yields:

$Cataphoric: Subject = 1$
$Anaphoric: Absence\ of\ Subject = 1$ (17)

*2) Features based on repetition rules:*
Feature values are filled as follows:

If я $(C_{Cat}, C_{Ana})$ is true then
  $Cataphoric: C_{Cat} = 1$
  $Anaphoric: C_{Ana} = 1$ (18)

In the example, considering EDU1 and EDU3 yields:

$Cataphoric: Object = 1$ (19)

$Anaphoric: Subject = 1$

*3) Features based on addition rules:*
Feature values are filled as follows:
If $Cataphoric:Д\ (Marker, After)$ or $AnaphoricД:(Marker, Before)$ is true then
  $Cataphoric: Marker\ After = Anaphoric: Marker\ Before = Marker$ (20)

In the example, considering EDU1 and EDU2 yields:

$Cataphoric: Marker\ After = Anaphoric: Marker\ Before = $ "และ" *(and)*

However, if one side of the pair is a relation, only addition and repetition rules are considered.

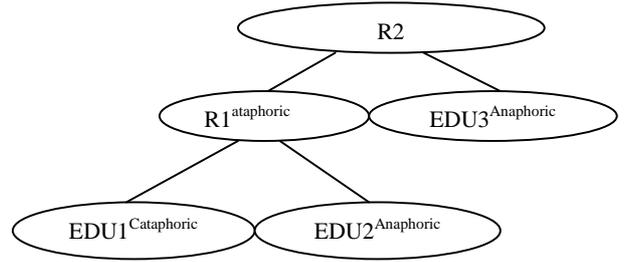

Fig. 7. Discourse relations in a rhetorical structure tree.

## VII. EXPERIMENTAL EVALUATION

### A. Evaluation of Thai EDU Segmentation

In order to evaluate the effectiveness of the EDU segmentation process, a consensus of five linguists, manually segmenting EDUs of Thai family law, is used. The dataset consists of 10,568 EDUs in total.

The EDU segmentation model is trained with 8,000 random EDUs, and the rest are used to measure performance.

The training continues until the estimated transition probability changes no more than a predetermined value of 0.02, or the accuracy achieves 98%.

The performances of both phrase identification and EDU segmentation are evaluated using recall (Eq. 21) and precision (Eq. 22) measures, which are widely used to measure performance.

$$\mathrm{Re}\,call = \frac{\#correct\ (phrases\ or\ EDUs)\ identified\ by\ HMM}{\#(phrase\ or\ EDUs)\ identified\ by\ linguists} \quad (21)$$

$$\Pr ecision = \frac{\# correct\ (phrases\ or\ EDUs)\ identified\ by\ HMM}{total\ \#\ (phrases\ or\ EDUs)\ identified\ by\ HMM} \quad (22)$$

The results show that the proposed method achieves the recall values of 84.8% and 85.3%; and the precision values of 93.5% and 94.2% for phrase identification and EDU segmentation, respectively.

*B. Evaluation of EDU Constituent Grouping*

In order to evaluate the effectiveness of the EDU constituent grouping, three corpuses are used which consist of Absence data (84 EDUs), Repetition data (117 EDUs) and a subset of the Family law with 367 EDUs). The Absence data contains EDUs mostly those following the Absence rules while the Repetition data contains mostly those following the Repetition rules. Five linguists create training and testing data sets by manually grouping EDU constituents.

Table 5 shows the results of grouping EDU constituents (subject (S), object (O), indirect object (I) and nomen (N)) by using rules based on NPs, assuming the positions of verb phrases (Vi, Vt and Vtt) are known. From the results, in general all rules, except $NP_O\text{-}NP_S\text{-}Vtt\text{-}NP_I$ and $NP_I\text{-}NP_S\text{-}Vtt\text{-}NP_O$, perform well.

Table 5: Performance of grouping EDU constituents

| Rules | Absence Data | Repetition Data | Family Law |
|---|---|---|---|
| $NP_S\text{-}Vi\text{-}NP_S$ | $NP_S$ (100%) | $NP_S$ (100%) | $NP_S$ (100%) |
| $NP_O\text{-}NP_S\text{-}Vt\text{-}NP_O$ | $NP_S$ & $NP_O$ (100%) | $NP_S$ & $NP_O$ (100%) | $NP_S$ & $NP_O$ (100%) |
| NPS-Vtt-NPO-NPI | $NP_S$ & $NP_O$ & $NP_I$ (100%) | $NP_S$ & $NP_O$ & $NP_I$ (100%) | $NP_S$ & $NP_O$ & $NP_I$ (100%) |
| $NP_O\text{-}NP_S\text{-}Vtt\text{-}NP_I$ $NP_I\text{-}NP_S\text{-}Vtt\text{-}NP_O$ | $NP_S$ (100%), $NP_O$ & $NP_I$ (91.37%) | $NP_S$ (100%), $NPO$ & $NP_I$ (79.59%) | $NP_S$ (100%), $NP_O$ & $NP_I$ (90.21%) |
| N-N | $NP_N$ (100%) | $NP_N$ (100%) | $NP_N$ (100%) |

To further resolve ambiguities with respect to these two rules, a probability table of terms in positions of $NP_I$ and $NP_O$ following Vtt ($P(Vtt| NP_I, NP_O)$) is used. The results of determining functions of EDU constituents by using the rules based on NPs together with the probability table show higher performance for Absence data (92.24%), Repetition data (85.78%), and Family law (93.71%).

*C. Evaluation of Thai RS Tree Construction*

In order to evaluate the effectiveness of the proposed Thai RS tree construction process, linguists manually construct the rhetorical structure trees of three texts used above with a total of 568 EDUs. The algorithms are evaluated by using recall (Eq. 23) and precision (Eq. 24) measures. Recall and precision are calculated with respect to how close an RS tree constructed from the proposed technique to that created by a consensus of the linguists.

$$\text{Re} call = \frac{\# correct\ internal\ nodes\ identified\ by\ RS\ Tree}{\#\ internal\ nodes\ identified\ by\ linguists} \quad (23)$$

$$\Pr ecision = \frac{\# correct\ internal\ nodes\ identified\ by\ RS\ Tree}{Total\ \#\ of\ internal\ nodes\ identified\ by\ RS\ Tree} \quad (24)$$

For the Absence and Repetition data sets, though relations between EDUs follow mostly Absence rules and Repetition rules, respectively, in reality when examined in details, many types of rules are used together in writing. For example,

Anaphoric EDU (S-Vt-O) : บุรุษไปรษณีย์ (S) จะคัดเลือก (Vt) จดหมาย (я O) (A Postman will sort letters)

Cataphoric EDU ((S)-Vt-O) : และ (Д) (Ф S) รับส่ง (Vt) จดหมาย (я O) (And will deliver letters)

Table 6 shows calculations of recall and precision of RS trees created by the Minimum Variance and Unweighted Arithmetic Average algorithms, in Fig. 8.

Table 6: RS tree construction performance of two clustering algorithms

| The correct RS tree | Minimum Variance | Unweighted Arithmetic Average |
|---|---|---|
| 3' | 3' | 3' |
| 4' | 4' | 4' |
| 1' | 1' | 1' |
| 9' | 9' | 6' |
| 2' | 2' | 2' |
| 5' | 5' | 5' |
| 6' | 6' | |
| 7' | 7' | |
| 8' | 8' | |
| | | 7' |
| | | 8' |
| | | 9' |
| | | 10' |
| | Precision = 9/9 | Precision = 6/10 |
| | Recall = 9/9 | Recall = 6/9 |

Table 7 shows the results of evaluating Thai RS Tree construction on the three data sets. The performance on the Family law dataset which combines many kinds of rules in its content is 94.90% recall and 95.21% precision. The results also show that Unweighted Arithmetic Average clustering algorithm gives the best performance for Thai RS Tree construction.

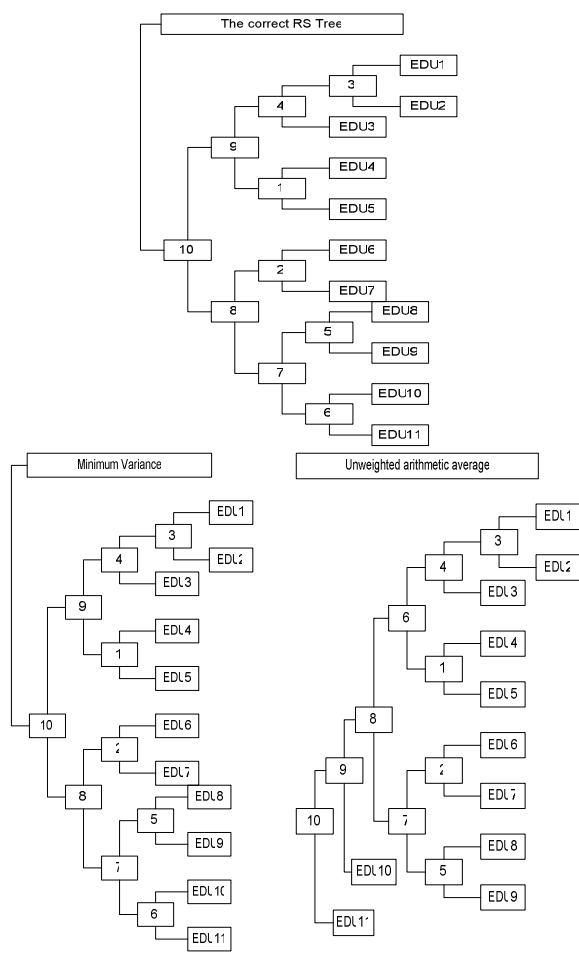

Fig. 8. RS trees from two hierarchical clustering algorithms

*D. Evaluation of Thai Discourse Relation Determination*

In order to evaluate the effectiveness of the Thai DR determination, linguists manually tag a relation to each internal node of RS trees constructed from Family Law, with a total of 624 EDU/relation pairs. A C5.0 decision tree algorithm [11] is trained with 424 random pairs, and the rest are used to measure performance. Ten discourse relations are studied in this research. Since markers are found helpful in determining relations, the test set is divided into EDU/relation pairs with markers and those without markers. The performance is reported accordingly.

Table 8 shows the results of determining ten discourse relations. The performance of EDU pairs from the Family law with and without marker is 85.09% and 82.81%, respectively.

After training with C5.0, some features are pruned, and the remaining features consists of: Subject, Object, Preposition, Nucleus, Modifier Nucleus, and Marker After for the cataphoric EDU, and Subject, Absence of Subject, Object, Absence of Object, Preposition, Absence of Preposition, Nucleus, Modifier Nucleus, and Marker Before for the anaphoric EDU.

Table 7: Performance of the RS tree construction

| Data | Num EDUs | Clustering Method | Recall | Precision |
|---|---|---|---|---|
| Absence | 84 | Neighbor Joining | 87.23 | 89.13 |
| | | Single Linkage | 82.97 | 84.78 |
| | | Un weighted Arithmetic Average | 87.23 | 89.13 |
| | | Minimum Variance | 89.40 | 91.30 |
| | | Weighted Arithmetic Average | 87.23 | 89.13 |
| Repetition | 117 | Neighbor Joining | 89.70 | 91.04 |
| | | Single Linkage | 83.82 | 85.07 |
| | | Unweighted Arithmetic Average | 89.70 | 91.04 |
| | | Minimum Variance | 77.94 | 79.10 |
| | | Weighted Arithmetic Average | 89.70 | 91.04 |
| Family-Law | 367 | Neighbor Joining | 85.98 | 86.26 |
| | | Single Linkage | 64.01 | 64.21 |
| | | Unweighted Arithmetic Average | 94.90 | 95.21 |
| | | Minimum Variance | 63.37 | 63.57 |
| | | Weighted Arithmetic Average | 90.44 | 90.73 |

Table 8: Performance of DR determination.

| Discourse Relations | Accuracy (%) | |
|---|---|---|
| | Without Marker | With Marker |
| คล้อยตาม (consent) | 93.10% | 98.10% |
| ตัวอย่าง (example) | 52.40% | 54.00% |
| ลักษณะวิธี (characteristic) | 69.40% | 99.30% |
| สรุปความ (summary) | 96.10% | None |
| เงื่อนไข (condition) | 59.60% | 85.30% |
| เลือกเอา (option) | 97.70% | 99.40% |
| เวลา (time) | 62.50% | 90.50% |
| เหตุผล (reason) | 90.80% | 91.20% |
| แจกแจง (explanation) | 100.00% | None |
| แตกต่าง (contrast) | 92.00% | 98.90% |
| Overall | 82.21% | 85.09% |

According to a sensitivity analysis, Marker ranks at the top for determining discourse relations. This also shows in the results where the accuracy of determining relations for EDU/relation pairs with markers is higher than for those without markers.

VIII. CONCLUSIONS

Rhetorical structure analysis explores relations among elementary discourse units (EDUs) in a text. It is very useful for many textual analysis applications such as automatic text summarization and question-answering.

This article proposes a novel technique to analyze rhetorical structure of Thai texts which combines machine learning techniques with linguistic properties of Thai language. Relations among EDUs are expressed hierarchically as a rhetorical structure tree.

First, phrases are determined and then are used to segment EDUs. The phrase segmentation model is a hidden Markov model constructed from the possible arrangements of Thai phrases based on part-of-speech of words, and the EDU segmentation model is a hidden Markov model constructed from the possible phrase-level arrangements of Thai EDUs. Linguistic rules are applied after the EDU segmentation to group related constituents into large units. Experiments show the EDU segmentation effectiveness of 85.3% and 94.2% in recall and precision, respectively.

A hierarchical clustering algorithm whose similarity measure derived from semantic rules of Thai language is then used to construct an RS tree. The technique is experimentally evaluated, and the effectiveness achieved is 94.90% and 95.21% in recall and precision, respectively.

Once an RS tree is constructed, a decision tree algorithm whose features derived from the semantic rules is used to determine discourse relations between EDUs in the tree. The technique is experimentally evaluated, and the overall effectiveness is at 82.81%.

AUTHORS PROFILE

**Somnuk Sinthupoun** is an instructor at the Department of Computer Science, Maejo University, Thailand. He holds an M.S. in Computer Science from National Institute of Development Administration (NIDA.), and a B.S. in Computer Science from Maejo University. His main research interests include artificial intelligence, information retrieval, data mining, and related areas.

**Ohm Sornil** is an Assistant Professor at the Department of Computer Science, National Institute of Development Administration, Thailand. He holds a Ph.D. in Computer Science from Virginia Polytechnic Institute and State University, an M.S. in Computer Science from Syracuse University, an M.B.A. in Finance from Mahidol University, and a B.Eng. in Electrical Engineering from Kasetsart University. His main research interests include computer and network security, artificial intelligence, information retrieval, data mining, and related areas.